\documentclass[letterpaper, 10 pt, conference]{ieeeconf}  

\IEEEoverridecommandlockouts                              
\usepackage{graphicx}
\usepackage{amsmath}

\graphicspath{ {images/} }

\title{
A Problem Reduction Approach for Visual Relationships Detection
}

\author{Toshiyuki Fukuzawa \\%
\texttt{t.fukuzawa@gmail.com} \\
\\
European Conference on Computer Vision 2018 Workshop (ECCV 2018)
}

\begin{document}

\maketitle
\thispagestyle{empty}
\pagestyle{empty}

\begin{abstract}

Identifying different objects (man and cup) is an important problem on its own, but identifying the relationship between them (holding) is critical for many real world use cases. This paper describes an approach to reduce a visual relationship detection problem to object detection problems. The method was applied to Google AI Open Images V4 Visual Relationship Track Challenge \cite{eccv_challenge}, which was held in conjunction with 2018 European Conference on Computer Vision (ECCV 2018) \cite{eccv} and it finished as a prize winner. The challenge was to build an algorithm that detects pairs of objects in particular relations: things like "woman playing guitar," "beer on table," or "dog inside car.". The dataset includes both object bounding boxes and visual relationship annotations. The training set contains 1.7 million images with 3 million bounding box annotations for 329 distinct relationship triplets, occurring a total of 374,768 times. This Open Images V4 dataset follows in the tradition of PASCAL VOC \cite{pascal}, ImageNet \cite{imagenet} and COCO \cite{coco}, but at an unprecedented scale.

\begin{figure}[thpb]
      \centering
      \includegraphics[width=40mm]{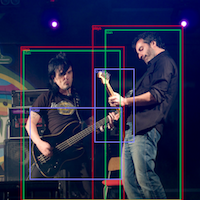}
      \includegraphics[width=40mm]{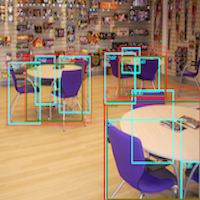}
      \caption{Examples of object relationship "man playing guitar" and visual relationship "table is wooden". Left: Mark Paul Gosselaar plays the guitar by Rhys A. Right: Civilization by Paul D.}
      \label{figurelabel}
   \end{figure}

\end{abstract}

\section{Introduction}

\subsection{Open Images V4 Dataset and Challenge}

Open Images V4 Dataset is one of the largest image recognition benchmark dataset of 1.9 million images with object location annotations. The bounding boxes are largely manually drawn by professional annotators to ensure accuracy and consistency. The images are very diverse and often contain complex scenes with several objects. The dataset size is about 600 giga bytes and the annotation labels of 1.7+ million images were provided as the training set for the challenge. 

The Visual Relationships Detection challenge requires detecting relationships connecting two objects. These include both human-object relationships (e.g. "woman playing guitar", "man holding microphone") and object-object relationships (e.g. "beer on table", "dog inside car"). Each relationship connects different pairs of objects, e.g. "woman playing guitar","man playing drum". Finally, this track also consider object-attribute relationships (e.g."handbag is made of leather" and "bench is wooden").

In the notation, a pair of objects connected by a relationship forms a triplet (e.g. "beer on table"). Visual attributes are in fact also triplets, where an object in connected with an attribute using the relationship is (e.g. "table is wooden"). The annotations are based on the image-level labels and bounding box annotations of Open Images V4. 467 possible triplets were initially selected and annotated on the training set of Open Images V4. The 329 of them that have at least one instance in the training set formed the final set of triplets for the Visual Relationships Detection track. It involves 62 different object classes.

\begin{table}[h]
\caption{ Number of Annotations in Training Dataset }
\label{table_distribution2}
\begin{center}
\begin{tabular}{|c||c||c|}
\hline
Relationship Triplets  & Bounding Boxes & Image-level Labels \\
\hline
374,768	  & 3,290,070	 & 2,077,154 \\
\hline
\end{tabular}
\end{center}
\end{table}

\begin{table}[h]
\caption{Number of Distinct Classes and Triplets}
\label{table_distribution}
\begin{center}
\begin{tabular}{|c||c||c||c|}
\hline
Type  & Classes & Relationships & Triplets \\
\hline
is & 23 & 1 & 42\\
\hline
non-is & 57 & 9 & 287 \\
\hline
total & 62 & 10 & 329 \\
\hline
\end{tabular}
\end{center}
\end{table}

\subsection{Performance Evaluation}

Model performances are evaluated by computing the weighted mean of three metrics.

\begin{itemize}

\item Mean Average Precision of relationships detection at IoU $>$ threshold ($mAP_{rel}$).
\item Recall@N of relationships detection at IoU $>$ threshold (Recall@$N_{rel}$).
\item Mean Average Precision of phrase detection at IoU $>$ threshold ($mAP_{phrase}$).

\end{itemize}

where Intersection-over-Union (IoU) threshold = 0.5. The weights applied to each of the 3 metrics are [0.4, 0.2, 0.4]. More details of the evaluation metrics and the evaluation server code are available online \cite{c_eval}\cite{c_eval2}.

\section{Method}

\subsection{VRD Problem Reduction to Object Detection Problems}

The main idea of the approach is to reduce the visual relationships detection problem to two sub object detection problems and a relationship finding problem. With this approach, you can leverage more widely studied object detection problem approaches \cite{tenfpaper}\cite{mobilenet}\cite{mask} and frameworks, for example, Tensorflow Object Detection API \cite{tenf}. Since the challenge dataset contains 2 types of relationships - "is" visual relationship to detect a single bounding box with visual attribute and object  relationships which connect two objects, these 2 types were treated separately by different neural net models and concatenated at the end. Object detection approaches can be applied directly on the "is" visual relationships, by handling each visual relation triplet as a separate target class. Object relationships can be computed with Gradient Boost Decision Tree ( LightGBM ) \cite{lightgbm} \cite{lightgbm2} \cite{lightgbm3} based on the output of object detection outputs as its features. The final model was produced as the concatenation of the 2 models. 

\begin{figure}[thpb]
      \centering
      \includegraphics[width=70mm]{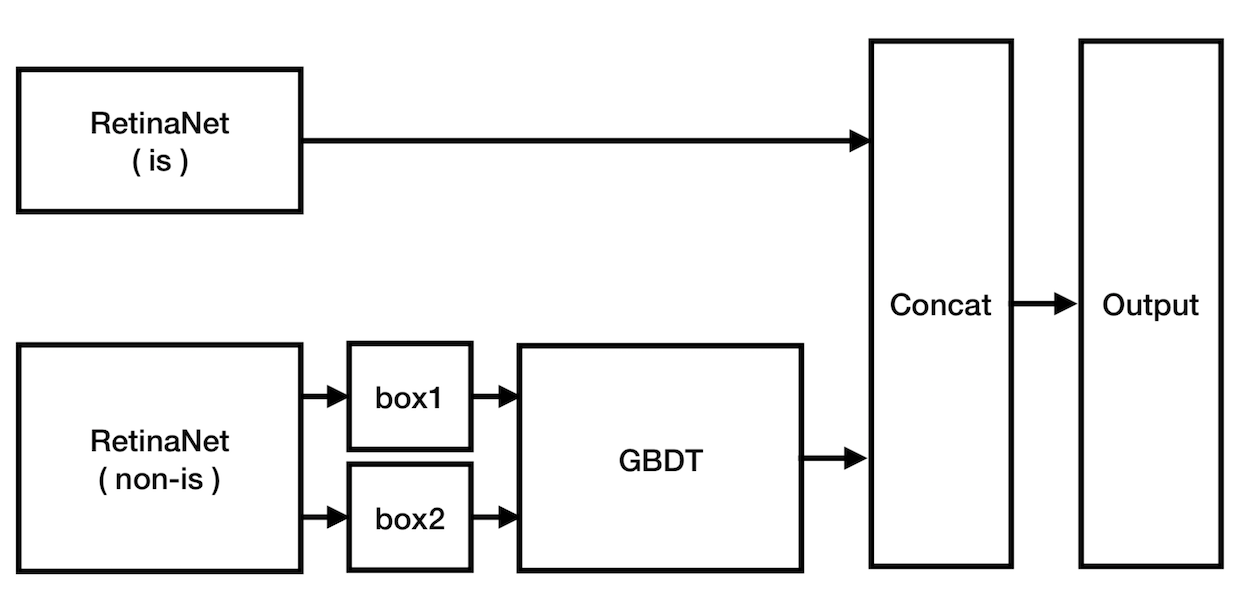}
      \caption{Final Model Architecture for VRD}
      \label{figurelabel}
   \end{figure}

\subsection{Object Detection with SSD-Resnet50-FPN (RetinaNet)}

RetinaNet ( SSD \cite{ssd} with Resnet50 \cite{resnet} backbone and Feature Pyramid Network (FPN) \cite{fpn} optimized on classification Focal loss \cite{retinanet} and localization loss ) was applied to both the "is" visual attribute and the bounding boxes for object relationships detection models. The backbone was trained on (640,640) fixed re-scaled inputs to optimize TPU training performance. The classification focal loss and localization loss weights were set to 1:1. The models were trained with SGD optimizer with momentum. The learning schedule was a cosine decay with base LR 0.04, warm-up LR 0.013 and warm-up steps 2000 with batch size 64. The total training steps were 25,000 for "is" visual relationships detection and 50,000 for bounding boxes detection for object relationships. The max number of boxes per image for each prediction was set to 100.

\begin{figure}[thpb]
      \centering
      \includegraphics[width=80mm]{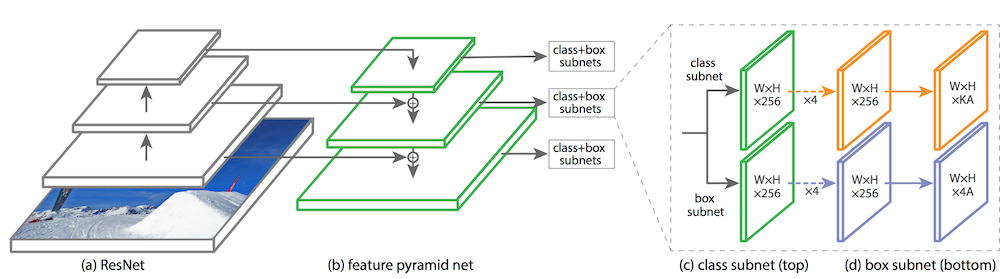}
      \caption{SSD Resnet FPN with Separate Classification Focal Loss and Localization Loss channels. This visualization is quoted from Tsung-Yi Lin et al's work \cite{retinanet}}
      \label{figurelabel}
   \end{figure}

\subsection{Class Imbalance and Focal Loss}

\begin{figure}[thpb]
      \centering
      \includegraphics[width=75mm]{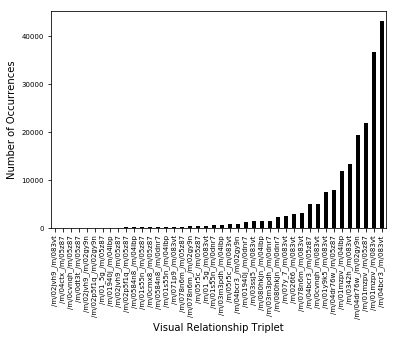}
      \caption{Visual Relationship Triplet Distribution}
      \label{fig_stats1}
   \end{figure}

\begin{figure}[thpb]
      \centering
      \includegraphics[width=75mm]{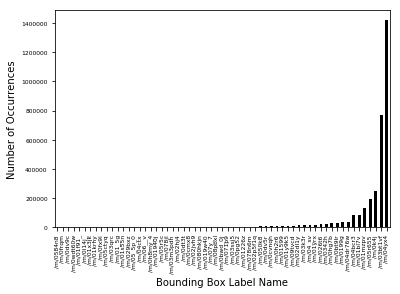}
      \caption{Bounding Box Class Distribution}
      \label{fig_stats2}
   \end{figure}

The significant imbalance of classes in the training dataset was a challenge for both "is" visual relationships and object relationships detections. As you see in the figure \ref{fig_stats1} and figure \ref{fig_stats2}, in the "is" relationship triplets, the lowest and highest frequent samples ratio is 1:8600 and in the object relationships bounding boxes, the ratio is 1:8700.

To deal with this extreme imbalance, the Focal Loss \cite{retinanet} was applied for the models to learn hard examples efficiently. 

\[ FL(p, y) =
  \begin{cases}
    -\alpha (1 - p)^{\gamma} \log{(p)}       & \quad \text{if } y = 1 \\
    -(1-\alpha) p^{\gamma} \log{(1-p)}  & \quad \text{otherwise}
  \end{cases}
\]

where $y \in \{0, 1\}$ specifies the ground truth class and $p \in [0, 1]$ is the model's estimated probability for the class. The gamma modulating factor $(1-p)^{\gamma}$ adjusts the cross entropy loss to down weight easy examples and thus allow the model to focus training on hard examples. The alpha constant factor balances the importance of positive and negative examples. Both the models were trained with $\gamma = 2, \alpha = 0.25$ . With $\gamma = 2$, an example classified with p = 0.9, y = 1 would have 100 times lower loss than the simple cross entropy of $\gamma = 0$. 

\subsection{“IS” Visual Relationships}

The number of distinct triplets of “is” relationships was only 42, therefore each triplet was handled as a separate class. In the given triplet labels challenge-2018-train-vrd.csv, there were 194K “is” relations out of 374K samples. The training set for the neural net was generated with these triplets for the positive labels, adding 10K images which don’t have any “is” relations as negative labels. This single model performed at metrics score 0.09 on the test dataset.

\subsection{Relationships Connecting Two Objects}

The neural net model was trained on challenge-2018-train-vrd-bbox.csv positive labels with down sampled negative 150K images to detect bounding boxes. Then, in each image, the top 100 valid box combinations with the highest confidence out of 10,000 combinations ( 100 boxes $\times$ 100 boxes ) were chosen by GBDT. The confidence of the combination $C_c$ was given by

$$
C_c = F_r \sqrt{C_{box1} C_{box2}} 
$$

where $C_{box1}$ and $C_{box2}$ are the confidences for each box given by the box detection neural net model. $F_r$ is the relationship coefficient function for each box1-box2-relationship combination. Features of box labels, relationship label, Euclidean distance of boxes, relative Euclidean distance ( distance divided by sum of total box area ), relative x-y position of box1 to box2 and raw box coordinates were given to the GBDT model. The GBDT model was trained with cross entropy loss with num\_leaves = 31, learning\_rate = 0.05, feature\_fraction = 0.9, bagging\_fraction = 0.8 and bagging\_freq = 5. This single model performed at metrics score 0.16 on the test dataset.

\subsection{Resources}

The neural net models' training was performed on Google Cloud Tensor Processing Unit (TPU) with 8 shards, which allowed to set a large batch size for each step and process training data on Google Storage efficiently. The model evaluation and inference were performed in parallel on a Tesla V100 GPU. Each training finished within 12 hours for 10 epochs of the training dataset. The GBDT training was far quicker and performed on a 16 CPUs machine.

\subsection{Ensemble}

The final model is obtained by just concatenating the first and second model outputs. Since they detect different types of relationships, there was no conflict or performance degradation. This final model performed at metrics score 0.25 on the test dataset.

\section{Conclusions}

This paper showed an approach to reduce a visual relationships detection problem to object detection problems which are solvable by more commonly available neural network architectures for object detection. An application of the approach was competitive in the Open Images V4 challenge and was awarded the prize in the Visual Relationships Detection track.

\addtolength{\textheight}{-12cm}   




\section*{ACKNOWLEDGMENT}

Special thanks to Google AI for making such a large and high quality dataset of Open Images V4 publicly available for machine learning experiments.


\end{document}